# SIMULATING BRAIN REACTION TO METHAMPHETAMINE REGARDING CONSUMER PERSONALITY

Maryam Keyvanara[1] and Seyed Amirhassan Monadjemi[2]

[1]Department of Computer Engineering, University of Isfahan, Isfahan, Iran
[2] Department of Computer Engineering, University of Isfahan, Isfahan, Iran

## ABSTRACT

*Addiction, as a nervous disease, can be analysed using mathematical modelling and computer simulations. In this paper, we use an existing mathematical model to predict and simulate human brain response to the consumption of a single dose of methamphetamine. The model is implemented and coded in Matlab. Three types of personalities including introverts, ambiverts and extroverts are studied. The parameters of the mathematical model are calibrated and optimized, according to psychological theories, using a real coded genetic algorithm. The simulations show significant correlation between people response to methamphetamine abuse and their personality. They also show that one of the causes of tendency to stimulants roots in consumers personality traits. The results can be used as a tool for reducing attitude towards addiction.*

## 1. INTRODUCTION

Drug epidemics are becoming one of the main concerns in countries all around the world. Addiction of drugs can be categorized into addiction to narcotics and stimulants [1, 2]. Unfortunately the use of stimulants has become very popular in the modern world and in particular in Iran. These stimulants, in general, include cocaine, ecstasy pills and crack. Likewise, one of the very dangerous stimulant substances is methamphetamine, also known as "crystal" or "ice". Methamphetamine, as a drug of abuse, has received a lot of attention in the recent years. It is an industrial, highly addictive substance. Different types of administration for methamphetamine include snorting, injecting, smoking and ingestion in the form of pills. Among these methods, injections gets the consumer to the "high" very quickly [1, 3].

Research shows that there is a relevance between tendency to stimulant consumption and the consumers' personality [4, 5]. Brain reaction to stimulants in terms of personality extraversion is of great importance. An individual's affinity to drug abuse according to introversion or extraversion of his/her personality is different. The approach to consumption of stimulants is much greater in extroverts compared to introverts [4-6]. Previous research can show that consumers of these substances are in different personality categories and cannot show whether this difference is due to their personality or perhaps is relevant to variables that interact with their personality. In this study, only variables of personality and addiction are considered and in fact tens of other possible variables that may be consistent with personality can be controlled and a causal relationship between personality and addiction can be displayed [4-7]. The uniqueness of this study is that is uses methamphetamine, which is a widely spread substance in Iran, as the stimulant.

The aim of this research is to observe and analyse, by means of computer simulation, different responses of consumers with three personalities of introvert, ambivert, and extrovert to a single dose of a methamphetamine. Brain dynamics during stimulant consumption are demonstrated using an existing mathematical model. This model shows the effects of stimulant drugs on human brain [4]. The reason for these simulations is that the real evaluation of these effects on humans is not possible because of moral and legal considerations. Since methamphetamine is a dangerous substance, it cannot be tested on non addicts. To use the mathematical model, its parameters should be calibrated according to the opponent process theories for addiction provided in psychology [4]. This calibration is done using genetic algorithm which is an optimization evolutionary method. The suggested method is implemented and executed in MATLAB 2013 and examines several states and consumption patterns of drug abuse [8].

In the following sections, we will present a review on personality, neural system and the related theoretical concepts. The suggested method is described in sections 3 and 4 and the simulation results are presented in section 5.

## 2. PERSONALITY AND HUMAN NERVOUS SYSTEM

Human body, in particular the nervous system, and the brain, are very complicated in structure. Human brain weighs about 1.4 kilograms and is the most complex system ever known. This system consists of hundreds of millions of neurons that interact with each other in complex manner. Studying this system is an interdisciplinary topic and is not limited to just one scope of science [9]. This complexity affects many aspects of the body and mind including personality, the way people approach stimulants, the amount of consumption, and how they quit the consumption of drugs. Therefore, it is of great importance to have a comprehensive cognition on the nervous system in terms of personality which is the reason why many people are inclined to stimulants.

In psychology, different personality traits are defined for human beings. One of these traits is extraversion, which according to unique personality trait theory, is known as the fundamental personality trait [10-12]. The biological basis of this trait is brain general activation system. According to unique personality trait theory, anxiety and impulsivity are two ends of a trait called extraversion. Extraversion starts with anxiety which has introversion and low psychoticism at one end and goes on to the other end of the pole with impulsivity, extraversion, and high psychoticism [10-12]. Also the opponent process theory presented by Solomon and Corbit in 1974, defines the response to an emotional stimuli in two main variables [4, 13].

When a person consumes stimulants, personal traits have great impact on person's actions and brain activity. Every individual, based on personal differences between people and also personality variables, shows a different reaction to stimulant consumption. This difference in reaction can be an important issue for drug abuse in various people [4, 14]. Before a person starts the stimulant consumption, their extraversion defines their brain tonic activity. When the substance is consumed, their extraversion level is altered [4, 5, 12]. Moreover, since the stimulant changes brain activation level and consumers differ in terms of their tonic activation level, then a dynamic model that considers both of these factors can predict stimulants effects on people with different personalities [4].

Many natural and biological phenomena can be expressed using a system of delayed differential equations [8, 15]. Dynamic intelligent systems with capability of prediction, in general should be a good instance of the real world they are modelling and embrace, as much as possible, all the conditions of the event they are predicting [15, 16]. Human brain response to addiction and drug consumption is also something which can be modelled mathematically, for example using the General Modelling Methodology, and simulated numerically [4, 5, 17-19].

Many studies have introduced mathematical models for the brain to show its dynamics, specifically in cases of addiction [17-21]. Haken introduced a dynamic model called lighthouse which is a mathematical model that demonstrates the dynamics of the brain and works on the relationship between brain subsystems that are defined by the neurons [21]. Also, in 2008 Amigo et al presented a dynamic mathematical model to show the brain response to a single dose of cocaine hydrochloride [4]. In 2010 Caselles introduced a more comprehensive model which embraces addiction to stimulants and is more extensible [5]. Models presented for dynamic systems consist of parameters that should be initialized before using the system. For this purpose, genetic algorithm can be considered as a suitable optimizing method. This algorithm has been used for the calibration of water distribution model, calibration of hydraulic process model used for detection of water loss and many other parameter calibration problems [22-29]. Also for determining the optimum soil parameters, evolutionary algorithms have been used. Genetic algorithm and particle swarm optimization methods have shown good results in these optimization problems [30, 31]. Particle swarm optimization has also been used in the calibration of climate models and has optimized the model parameters [32]. Also, in a study done in 2008, a genetic algorithm has been used for the generation of the system of differential equations. This cognitive research, studies the existence of chaotic response of a person consuming a dose of stimulants [33].

## 3. MODELLING HUMAN RESPONSE TO STIMULANTS

The goal of this study is the analysis of the effect of a single dose of methamphetamine on consumers' personality. An existing mathematical model introduced in 2008 by Amigo et al is used for this purpose. This model is a set of delayed differential equations (DDE) and functions based on consumers' personality trait in a psychological perspective and also the brain characteristics [4]. The solution of these equations displays the consumer's brain activity level. In other words, by solving the equations for a given time interval, we can observe the changes in the individual's personality in terms of extraversion.

Having this mathematical model, we need to implement a computer program in Matlab, based on the model so that the simulations are done correctly. Also, the presented model should be calibrated with correct parameter values. This calibration must be accomplished in a way that model variables have suitable parameters and the creation of the model and its usage by the user are not controllable. Moreover after the calibration, the created system should show a known behavior in the simulations. For using the mathematical model, settings of the parameters have to be carried out according to the pharmacological information for the methamphetamine. In oral administration of methamphetamine, peak concentrations are seen in 2-3 hours after ingestion and reaches its maximum. This drug remains in the plasma for 4 to 12 hours [34-36].

In this model, a state variable, s(t), is defined which shows the amount of drug consumed in the blood at each instance t. This variable produces consumer's acute response to stimulant. The stimulant drug shows itself in the model as the amount of drug in blood and is defined as Equation (1).

$$s(t) = \frac{\alpha.d(e^{-\alpha t} - e^{-\beta t})}{\beta - \alpha} \quad , \qquad \alpha \neq \beta \tag{1}$$

International Journal of Artificial Intelligence & Applications (IJAIA) Vol. 6, No. 4, July 2015

In this equation, d is the dose of the consumed stimulant and α and β are assimilation and consumption rates of methamphetamine in the body and their values must be determined before simulations. As mentioned before, the dose of the consumed drug is one dose of methamphetamine and so d equals one. As for α and β, they should be determined in a way that the maximum point in s(t) happens when the drug level reaches its highest degree of concentration in the blood. This point happens after 2-3 hours of drug entrance to the body [34-36]. In the simulations we consider this time 120 minutes. Moreover, as we would like to observe the brain response for the duration of the time drug remains in the blood, the total time of simulation is considered 725 minutes and the start of intake is $t_0$=5 minutes. Before $t_0$, the consumer has no drug inserted to his body. Also a delay parameter of τ, which is the inhibitor effect delay, is set to 340 minutes.

To determine the values of α and β, a genetic algorithm is designed to find the optimum values of these controlling rates. A genetic algorithm with two populations is implemented in Matlab; a population for α and another for β. The fitness function is defined according to Equation (1) and setting t equal to 120, as presented in Equation (2).

$$s(120) = \frac{\alpha.d(e^{-\alpha*120} - e^{-\beta*120})}{\beta - \alpha}, \quad \alpha \neq \beta \qquad (2)$$

In genetic algorithm each gene is a candidate as the solution of the problem. Therefore the representation of the genes is important. In this optimization, the genes are floating point numbers between 0 and 1. Hence real number representation has been used in the implementation of the genetic algorithm [37, 38].

The proposed algorithm is initialized with generating random numbers between 0 and 1 for each of the two populations. To begin, the fitness of the generated numbers is calculated for each gene. For selection, the roulette wheel method is chosen. Roulette wheel selects parents directly proportional to their fitness. For each individual, the probability of being selected for the next generation is the fitness of that individual divided by the sum of the fitness values of the entire population. This selection method is applied for both populations [38].

The performance of genetic operators, crossover and mutation, in real coded GAs are essentially different from those in binary, while the drive and the framework is similar. In real GA, the operation of crossover considers each chromosome a real number and uses a determined crossover rate to swap the digits. In this method, one point crossover is applied [38]. Also, as the chromosomes of the population are real numbers, real mutation should be applied. Real number mutation differs with binary mutation. In it, a very small amount from a known distribution is added to each variable. In binary populations, mutation reverses each bit i.e. it changes zero to one and one to zero. Mutation operator includes adding a small amount from a uniform distribution in the interval of [-0.1,0.1] to a random place in the gene. Both populations go on to maximizing s(t) function and increase its value [39]. The criteria for the termination of the algorithm is that either the answers do not change over several successive generations, or that the algorithm is repeated for the determined number of generations. The values obtained for the parameters are shown in Table (1).

Table 1. Optimizing the Parameters Using Genetic Algorithm

| A | β | s(120) |
|---|---|--------|
| 0.0121 | 0.0071 | 0.4351 |

To avoid being trapped in local optima, the crossover and mutation rates must be selected with care. In this study, these values are obtained using trial and error. So, the values of 0.75 and 0.15 were selected respectively. If the mutation rate is too small, it will lead to premature convergence to a local optimum instead of the global one. On the other hand, if a very large value is selected, it will lead to missing good solutions during the search [39, 40].

The type of personality in individuals is expressed using the delayed differential equation in (3). The solution of this equation shows the extraversion of the consumer and needs to be solved using numerical methods.

$$\frac{dy(t)}{dt} = \begin{cases} a(b-y(t)) + \frac{p}{b}.s(t-t_0) & t_0 \leq t \leq t_0 + \tau \\ a(b-y(t)) + \frac{p}{b}.s(t-t_0) - b.q.s(t-\tau-t_0).y(t-\tau-t_0) & t \geq \tau \end{cases} \quad (3)$$

$$y(t_0) = y_0$$

In the delay differential equation mathematical model in (3), the tonic activation level of each individual is expressed with a constant parameter $b$. If the consumer has the basal personality of extraversion, the parameter b will be valued 0.5, for the introvert personality it would be 1.5 and for the ambivert this parameter would be 1. The consumer starts the consumption of a dose of stimulant with an initial extraversion value of $y_0$ which has one of the mentioned values. In fact it should be equal to the tonic activation level. This means that $y(t_0)=y_0=b$ [4].

As the consumer starts the administration of the stimulant, the extraversion parameter, defined by y, begins to be modified. Parameter $a$ is the homeostatic control rate and is a factor that causes the model to behave like the predicted form. The parameters $p$ and $q$ are excitation effect power and inhibition effect power. These values should be positive and depend on each individual. In the calibration of the model for methamphetamine, the value of parameter $a$ is set to 0.025, $p$ is 0.8 and $q$ is 0.15. Solving the equations in (3) analytically is very complicated. Hence the solution of these equations has been obtained using numerical methods for solving delayed differential equations (DDE) in MATLAB. These functions include dde23 and ddeset [8, 41]. The time interval for the simulations is equal to the duration of the time that the individual consumes one dose d of snorted methamphetamine until the effect of the drug reaches its highest degree of concentration in the blood and its duration in the blood ends. During the time methamphetamine remains in the blood, it is evident from the known psychological theories how the behaviour of the nervous system should be.

Model validation can be accomplished according to the time patterns produced in the simulations. This time pattern has to be consistent with the predicted behaviour predicted by the psychological theories. In fact the plots resulted from the simulations should show the unique personality trait theory for excitation-inhibition balance which is similar to affective acquisition pattern presented by Solomon and Corbit [12]. This all means that after the consumption of the stimulant, there should be an initial euphoria, then a feeling of pleasure which remains for some time. At the end, a displeasing activation is observed which causes an unpleasant feeling in the consumer. These are observable in the output results of the simulations.

## 4. SIMULATING BRAIN RESPONSE TO METHAMPHETAMINE

In this study, several simulations were conducted to observe the response of human brain to the methamphetamine stimulant according to the personality of consumer in terms of extraversion. Parameters of the model were calibrated with regard to this information, using genetic algorithms. Parameters must be determined so that the proposed time pattern for the model of the excitation-inhibition effect matches the time pattern predicted by Solomon and Corbit [12]. Since the presented model is a system of delayed differential equations, its solution must be found, for the expected time pattern, using numerical methods and MATLAB programming. Figure (1) shows the level of methamphetamine in body for the duration of 725 minutes.

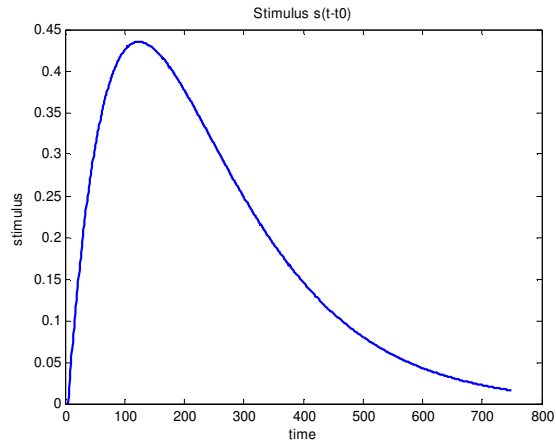

Figure 1. Stimulant level in the body after the consumption of a dose of methamphetamine

Parameters in these simulations are set in a way that their maximum point is 120 minutes (2 hours) after the start of consumption. The values of the parameters are found by genetic algorithm as explained in the previous section. According to Figure (1), as the time goes by, the concentration of stimulant in the blood increases until it reaches the maximum point of blood concentration. Then it starts to decrease as it is absorbed in the body. The stimulant level reaches values close to zero after 12 hours. The level of stimulant in the blood does not depends on the person's personality, but it depends on the dose of the consumed drug.

After the consumer snorts a dose of methamphetamine, he reaches a pleasant feeling. Then they experience a constant period of joy until the effect of drug starts to disappear and the level of activation makes the consumer to feel unpleasant. This part is shown in the negative part of Figure (2). At the end, it returns to the initial activation state. This figure shows the excitation-effect effect where the positive part illustrates the excitation factor and the negative part of it, is for the inhibitor.

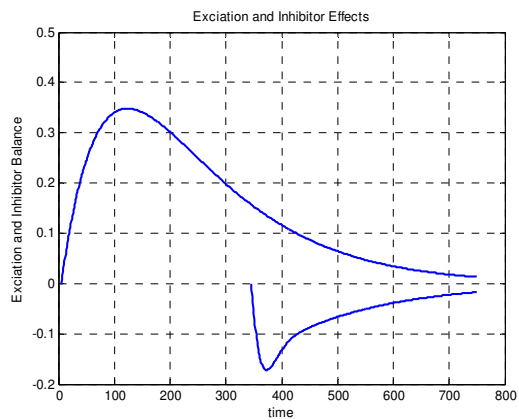

Figure 2. The Excitation and Inhibitor Effects of Methamphetamine

As it is observed in figure (2), the effect of inhibitor starts according to the value of the delay parameter at t=340 minutes. This is close to the time that the effect of drug is starting to be

International Journal of Artificial Intelligence & Applications (IJAIA) Vol. 6, No. 4, July 2015

reduced and eliminated. If we subtract the values of the inhibitor effect (negative part) from the excitation effect (positive part) we reach the plot in figure (3). The simulations in figure (3) show the excitation-inhibitor balance for three personalities of introvert, ambiverts and extroverts. The extrovert has been affected more than the other personalities by the excitation effect. As for the introvert, it is observable that he can return much faster to the tonic level (1.5 fo introvert) than the other personalities.

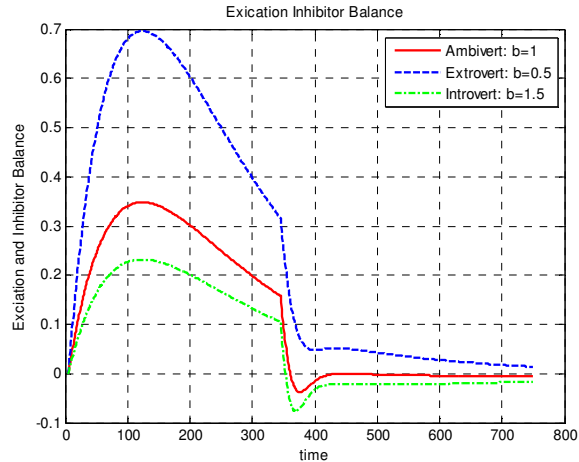

Figure 3. The excitation-inhibitor balance for introvert, ambivert and extroverts after the consumption of a single dose of methamphetamine

It is seen in figure (3) that there is stronger inhibitor effect in introvert than extrovert and the introvert has not had a very high excitation effect. The brain activation level that shows the extraversion of the consumers are shown in Figure (4). This simulation, shows the brain activation level for the duration of 725 minutes. The level of extraversion of the individuals increases until the concentration of the stimulant in the blood reaches its maximum point and then it starts to decrease. At the end, it returns to basic state. Figure (4) shows the level of extraversion for three types of personalities. The amount of tonic activation level (before any consumption) for the introvert is 1.5, for ambivert 1, and for the extrovert it is equal to 0.5.

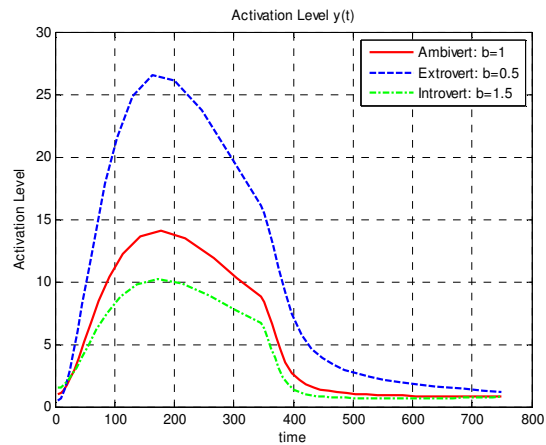

Figure 4. Extraversion variable for introverts, ambivert and extrovert after the consumption of a dose of methamphetamine

It can be observed that for the same period of time and for the same consumed dose, the extrovert has a higher brain activation level than the ambivert and the ambivert has bigger activation level than the introvert. The introvert returns to his tonic activation level much faster than the other personalities. The extrovert takes more time to return to his tonic level. This is due to the fact that extroverts have a higher brain excitation level. Introverts are not excited as much as the introverts by the dose of stimulant entered to their blood. Hence they have an easier and faster return to their basal brain activation level. Also, the increase in brain activation level, y(t), is very high compared to cocaine [4]. This is due to the nature of methamphetamine which a more powerful stimulant than cocaine.

## 5. CONCLUSION

With the acquisition of the system presented in this research which has the capability of predicting different responses of human brain to consumption of a single dose of methamphetamine, based on the type of extraversion of the individuals, we can develop new advances in addiction prevention. Evolutionary methods, in this study genetic algorithm, were used as a powerful tool to model parameter calibration and were found a great optimization method. Considering various consumption conditions, different consumption doses and stimulants other than methamphetamine, it can also help with prevention and treatment of addiction and drug abuse. By knowing that individuals who consume methamphetamine, have what kind of personality and based on this type of personality how their brains response to stimulants, there is a possibility for intervention and improvements of the addiction treatment. We observed that extroverts experience a much higher brain activation level than the introverts, after consuming one dose of methamphetamine. This intervention can be carried out on real human beings with the assurance of causality, and not relevance, and no other artifact variable has made this relevance. This study may underlie effective interventions to reduce the impact of addiction and lower the tendency to stimulants.

## 6. FUTURE SCOPE

Future work can use different, and at the same time popular, stimulant substances along with more data collected from the addicts consumption pattern. Moreover, since personality of human being has many traits, the models used in this research may be more generalized for different aspects of personality, not just extraversion, with correlation to stimulant consumption. On the other hand, the effects of other driving stimulus, like the effect of noise, can be analysed on introverts and extroverts.

## REFERENCES

[1] Baskin, I.S.D., Sommers, A.B., (2013) Methamphetamine Use, Personality Traits and High-Risk Behaviors, Springer.
[2] Thompson, K.M. H., Simon, S.L., Geaga, J.A., Hong, M.S., Sui, Y., (2004) "Structural Abnormalities in the Brains of Human Subjects Who Use Methamphetamine, Neurobiology of disease", Journal of neuroscience, vol. 24, pp. 6028- 6036.
[3] Roll, J. M., Rawson, R. A., Ling, W., & Shoptaw, S. (Eds.). (2009). Methamphetamine addiction: from basic science to treatment. Guilford Press.
[4] Amigo, S., Joan, A.C., Mico, C., (2008) "A Dynamic Extraversion Model. The Brain's Response to a Single Dose of a Stimulant Drug", British Journal of Mathematical and Statistical Psychology, 211-231.
[5] Caselles, A., Amigo, S., (2010), "Cocaine Addiction and Personality: A Mathematical Model", British Journal of Mathematical and Statistical Psychology, 449-480.

[6] Caselles, A., Micó, J. C., & Amigó, S. (2011). Dynamics of the General Factor of Personality in response to a single dose of caffeine. The Spanish journal of psychology, 14(02), 675-692.
[7] Hyman, R.C., (2001) "Addiction and the brain: The neurobiology of compulsion and its persistence", Nature Reviews Neuroscience, vol. 2, pp. 695- 703.
[8] Gieschke, R., Serafin, D., (2014) Development of Innovative Drugs via Modelling with MATLAB, A Practical Guide, Springer, 21-68.
[9] Davis, C. B., (2007) "Developing theory through simulation methods", Academy of Management Review, vol. 32, pp. 480- 499.
[10] Gray, J. A., (1972) "The psychophysiological nature of introversion-extroversion: A modification of Eysenck's theory, Biological basis on individual behavior", New York: Academic Press, 182–205.
[11] Gray, J. A., (1981) A critique of Eysenck's theory of personality, A model of personality, New York: Springer, 246-276.
[12] Amigo, S., (2005) "The unique-trait personality theory. Towards a unified theory of brain and conduct", Valencia: Editorial de la Universidad Politecnica de Valencia, 108-120.
[13] Solomon, R. L., Corbit, J. D., (1974) "An opponent-process theory of motivation. I. Temporal dynamics of affect", Psychological Review, 81, 119–145.
[14] Kirkpatrick, C.E.J., de Wit, H., (2013) "Personality and the acute subjective effects of d-amphetamine in humans", Journal of Psychopharmocology, vol. 27, pp. 256- 264.
[15] Allaire, A.C.G., (2007) Numerical analysis and optimization, An introduction to mathematical modelling and numerical simulation.
[16] Law, W.D.K., (2006) Simulation, Modelling, and Analysis, New York: McGraw - Hill.
[17] Caselles, A., (1994) "Improvements in the Systems-Based Models Generator SIGEM", Cybernetics and Systems: An International Journal, 81-103.
[18] Ahmed, M.G.S.H., Gutkin, B. (2009) "Computational Approaches to the Neurobiology of Drug Addiction", Pharmaco psychiatry.
[19] Ahmed, G.B., Gutkin, B.S., (2007) "The simulation of addiction: pharmacological and neuro-computational models of drug self-administration", Drug Alcohol Depend, vol. 8, pp. 304-311.
[20] Bobashev, G., Costenbader, E., & Gutkin, B. (2007). Comprehensive mathematical modeling in drug addiction sciences. Drug Alcohol Depend, 89(1), 102-106.
[21] Haken, H., (2008) Brain Dynamics, An Introduction to models and simulations, Springer.
[22] Wu, Y., Walski, T., Mankowski, G.H., Gurrieri, R., Tryby, M., (2002) "Calibrating Water Distribution Model via Genetic Algorithms", Venue in Proceedings of the AWWA IMTech Conference, April 16-19.
[23] Chuntian, Ch., Chunping, O., Chau, K.W., (2002) "Combining a fuzzy optimal model with a genetic algorithm to solve multi-objective rain fall-runoff model calibration", Journal of Hydrology, 268, 1-4, 72-86.
[24] Wu, P.S., (2006) "Water Loss Detection via Genetic Algorithm Optimization-Based Model Calibration", in ASCE 8th Annual International Symposium on Water Distribution System Analysis Cincinnati, Ohio.
[25] Wu, T.W., Mankowski, G.H., Gurrieri, R., Tryby, M., (2002) "Calibrating Water Distribution Model via Genetic Algorithms", Venue in Proceedings of the AWWA IMTech Conference, pp. 16- 19.
[26] Kumar, M.T.S., Raman, B., Sukavanam, N., (2008) "Stereo Camera Calibration using real Coded Genetic Algorithm", TENCON 2008-2008 IEEE Region 10 Conference.
[27] Mehrkanoon, S., Mehrkanoon, S., & Suykens, J. A. (2014). Parameter estimation of delay differential equations: an integration-free LS-SVM approach. Communications in Nonlinear Science and Numerical Simulation, 19(4), 830-841.
[28] Yazdi, F.K., Yazdi, H.S., (2012) "Calibration of Soil Model Parameters Using Particle Swarm Optimization", International Journal of Geomechanics, pp. 229- 238.
[29] Wu, Zh., Sage, P., (2006) "Water Loss Detection Via Genetic Algorithm Optimization-Based Model Calibration", ASCE 8th Annual International Symposium on Water Distribution System Analysis, Cincinnati, Ohio, August 27-30.
[30] Levasseur, S., (2008) "Soil parameter identification using a genetic algorithm", International Journal for Numerical and Analytical Methods in Geomechanics, vol. 32, pp. 189- 213.
[31] Espinosa, O.B., (2012) A Genetic Algorithm for the Calibration of a Micro-Simulation Model, arXiv.
[32] Kuok, C.P., (2012) "Particle Swarm Optimization for Calibrating and Optimizing Xinanjiang Model Parameters", International Journal of advanced Science and Applications, vol. 3.
[33] Caselles, S.A., (2008) "Chaos in Brain's Response to a single Dose of a Stimulant Drug", Systems Science European Union Congress Proceedings, System Complexity for Human Development.

[34] DiMaio, T. G., & DiMaio, V. J. (2005). Excited delirium syndrome: cause of death and prevention. CRC press.
[35] Barceloux, D. G. (2012). Medical toxicology of drug abuse: synthesized chemicals and psychoactive plants. John Wiley & sons.
[36] National Highway Traffic Safety administration: Drugs and Human performance Fact Sheets. (2015). Methamphetamine (and Amphetamine). Retrieved from http://www.nhtsa.gov/people/injury/research/job185drugs/methamphetamine.htm
[37] Rothlauf, F., (2006) Representation for Genetic and Evolutionary Algorithms, Springer, 2nd Ed, 10-25.
[38] Ashlock, D., (2006) "Evolutionary Computation for Modelling and Optimization", Springer, 66-74.
[39] Yoon,Y., Kim, Y., (2012) "The Roles of Crossover and Mutation in Real-Coded Genetic Algorithms", Bio Inspired Computational Algorithms and their Applications, Dr Shangse Gao (Ed.), 65-82.
[40] I. Zelinka, (2010) Evolutionary Algorithms and Chaotic Systems: Springer.
[41] Shampine, L.F.,Thompson, S., (2001) "Solving DDEs in MATLAB", Applied Numerical Mathematics, 37, 441-458.